\documentclass[11pt,letterpaper]{article}

\usepackage[margin=1in]{geometry}
\usepackage{setspace}

\usepackage[utf8]{inputenc}
\usepackage[T1]{fontenc}
\usepackage{amsmath}
\usepackage{newtxtext,newtxmath}
\usepackage{microtype}

\usepackage{graphicx}
\usepackage{caption}
\usepackage{booktabs}
\usepackage{tabularx}
\usepackage{array}
\usepackage{enumitem}

\usepackage{xcolor}
\usepackage{url}
\usepackage[natbibapa]{apacite}
\definecolor{linkblue}{rgb}{0.00,0.20,0.55}
\usepackage[colorlinks=true,linkcolor=linkblue,citecolor=linkblue,urlcolor=linkblue]{hyperref}
\hypersetup{
    pdftitle={Large Language Models for Injury Surveillance: Evidence from E-Bike Incidents},
    pdfauthor={Zhichao Yang, Faeze Momeni Rad, Jiashu He, Mohammad B. Al-Khasawneh, Darshan Pandit, Cinzia Cirillo}
}


\begin{document}

\title{\bfseries Large Language Models for Injury Surveillance:\\
Evidence from E-Bike Incidents}

\author{
Zhichao Yang$^{1}$, Faeze Momeni Rad$^{1}$, Jiashu He$^{2}$,\\
Mohammad B. Al-Khasawneh$^{1}$, Darshan Pandit$^{1}$, Cinzia Cirillo$^{1,*}$\\[0.8em]
{\small $^{1}$Department of Civil and Environmental Engineering, University of Maryland, College Park, USA}\\
{\small $^{2}$Department of Computer and Information Science, University of Pennsylvania, Philadelphia, USA}\\[0.8em]
{\small $^{*}$Corresponding author: Dr. Cinzia Cirillo}\\
{\small 1173 Martin Hall, College Park, MD 20742-6601, USA}\\
{\small Email: ccirillo@umd.edu}
}

\date{}

\maketitle

\begin{abstract}
This study analyzes e-bike-related injury incidents using two
complementary data sources: the Consumer Product Safety Risk
Management System (CPSRMS) and the National Electronic Injury
Surveillance System (NEISS). We develop an LLM-assisted framework
that converts unstructured incident narratives into structured
variables, notably human-, equipment-, and environment-related
contributing factors, for use alongside demographic, locational,
temporal, and weather-derived predictors.

To improve cross-source comparability, injury severity is
harmonized into three ordered categories. Separate ordered logit
models are estimated for CPSRMS and NEISS, together with two
supplementary joint scaled specifications. In CPSRMS, human-related
factors are associated with higher injury severity, whereas
equipment- and environment-related factors, riders under age 18,
spring incidents, and recreation/sport locations are associated
with lower severity categories. In NEISS, riders under age 18 and
female riders are associated with lower treatment severity, whereas
riders above age 59 are associated with higher severity.
Likelihood-ratio tests support selected coefficient-sharing
restrictions but reject equality of the source thresholds.

The findings demonstrate that narrative-rich incident reports
and standardized emergency-department surveillance data capture
complementary rather than directly interchangeable dimensions of
e-bike safety. The proposed framework illustrates how LLM-derived
variables can be integrated with econometric models to support
injury surveillance, safety intervention design, and evidence-based
micromobility policy.

\end{abstract}

\section{Introduction}

Electric bicycles (e-bikes) have experienced rapid growth in popularity
because of their convenience, cost-effectiveness, and environmental
benefits. In the United States, e-bike imports exceeded 2.2 million units
in 2025 \citep{LEVAssociation2025}, reflecting the continued expansion of
the domestic e-bike market.
E-bike participation has also increased substantially.
According to the Physical Activity Council, 19.4\% of
Americans who reported riding a bicycle in 2023
had used an e-bike \citep{PAC2023}.

This growth has been accompanied by increasing concern regarding
e-bike-related injuries and fatalities. The U.S. Consumer Product Safety Commission (CPSC) reported 53 deaths associated with e-bikes between 2017
and 2021 \citep{Tark2022}. E-bike injuries also accounted for
approximately 15\% of reported micromobility-related injuries during the
same general period \citep{CPSC2023micromobility}. These trends highlight
the need to better understand the human, mechanical, environmental, and
contextual factors associated with e-bike incidents.

Existing safety interventions commonly focus on protective equipment,
rider education, infrastructure, and vehicle technology. Helmet use, for
example, remains an important strategy for reducing the severity of head
injuries \citep{Ma2023HelmetEbike}. Technological developments may also
reduce risks associated with charging and battery failure. One such
system, SingMonitor, uses microphones embedded in mobile devices to
remotely assess the charging condition of e-bike power supplies
\citep{Jian2023SingMonitor}. Although these interventions address
specific risks, effective prevention also requires systematic evidence on
the mechanisms and circumstances associated with incidents and injury
severity.

A major obstacle to such analysis is the structure of available safety
data. Consumer reports, hospital records, and product-safety databases
often differ in their reporting mechanisms, severity definitions, and
level of contextual detail. Important information about rider behavior,
equipment malfunction, road conditions, and environmental circumstances
may appear only in free-text narratives. Manual review of large narrative
collections is resource intensive, difficult to standardize, and
challenging to scale across heterogeneous data sources.

Machine-learning and natural-language-processing methods provide tools
for addressing these challenges. Tree-based models, for example, have
been used to identify risk patterns among e-bike riders
\citep{Wang2019Entropy}. Large language models (LLMs) are particularly
well suited to extracting and organizing information from heterogeneous
text. In autonomous-system research, foundation models have been applied
to tasks including perception, reasoning, simulation, and planning
\citep{gao2024surveyfoundationmodelsautonomous}. Related advances in
domain adaptation, multistep reasoning, and multi-agent collaboration
have further expanded their analytical capabilities
\citep{lin2025sftdoesnthurtgeneral,xiong-etal-2024-large,xiong2024deliberate}.
LLM-based methods have also shown promise in biomedicine
\citetext{\citealp{he2024give}; \citealp{he2025selfgiveassociativethinkinglimited};
\citealp{wang2024surveylargelanguagemodels}}, recommendation systems
\citep{xurecommend,xurecommend2}, and misinformation analysis
\citep{xufake,xumulti,jin2025two}. Improvements in learning algorithms
and computational efficiency have made these approaches increasingly
practical for large-scale data processing
\citep{jin2024learning,yang2024hades}.

This study develops an LLM-assisted injury-surveillance framework for
e-bike incidents using two complementary CPSC data sources: the Consumer
Product Safety Risk Management System (CPSRMS) and the National Electronic
Injury Surveillance System (NEISS). CPSRMS contains
relatively detailed incident reports and narratives submitted by
consumers, manufacturers, and other authorized reporters, whereas NEISS consists
primarily of standardized emergency-department records supplemented by
short clinical narratives. Their contrasting structures provide an
opportunity to examine how narrative-rich reports and structured
surveillance records capture different dimensions of injury risk.

\begin{figure}[!tbp]
    \centering
    \includegraphics[width=0.88\linewidth]{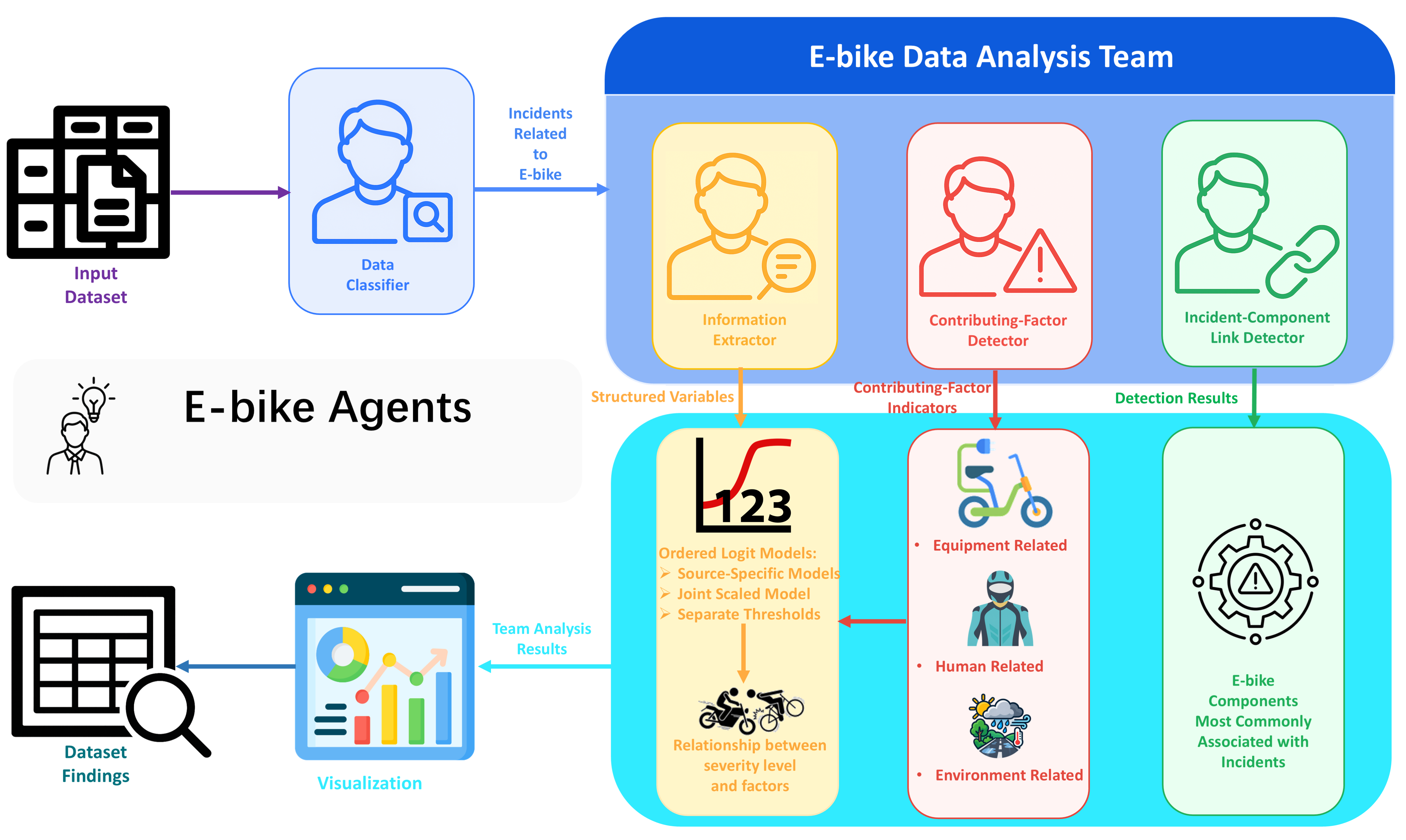}
    \caption{%
    Overall architecture of the LLM-assisted e-bike
    injury-surveillance framework, including case
    classification, information extraction,
    contributing-factor and component detection,
    and source-specific and joint scaled ordered
    logit severity modeling.}
    \label{fig::architecture}
\end{figure}

As illustrated in Figure~\ref{fig::architecture}, the framework uses
specialized LLM agents to identify e-bike-related records,
extract demographic and contextual information, classify contributing
factors, and determine whether specific mechanical or visibility
components are associated with an incident. Narrative information is
converted into structured indicators of human-, equipment-, and
environment-related factors that can be combined with conventional
demographic and locational variables.

To support cross-source injury-severity analysis, the original
source-specific severity outcomes are harmonized into three ordered
categories. We estimate separate ordered logit models for CPSRMS and NEISS and use two joint scaled ordered logit specifications as supplementary assessments of cross-source comparability.

The study makes three principal contributions. First, it demonstrates how
LLMs can convert heterogeneous incident narratives into structured
variables suitable for econometric analysis. Second, it develops a comparative modeling framework that prioritizes source-specific ordered logit models while using two joint scaled specifications and likelihood-ratio tests to assess cross-source comparability. The tests support selected coefficient-sharing restrictions but reject equality of the source-specific thresholds, indicating that CPSRMS and NEISS should not be treated as fully interchangeable. Third, it provides empirical
evidence that the two sources capture complementary dimensions of injury
severity: narrative-derived behavioral, equipment-related, and
environment-related factors are more prominent in CPSRMS, whereas
demographic characteristics are more consistently associated with
treatment severity in NEISS. Together, these contributions support more
comprehensive injury surveillance and provide evidence for behavioral
interventions, equipment oversight, and demographic-sensitive
micromobility safety policies.

\section{Related Work}

Existing e-bike safety research has examined rider characteristics
and behavior, vehicle-related risks, and roadway and environmental
conditions. These dimensions jointly influence incident occurrence
and injury severity, while comparisons across transportation modes
highlight differences in rider populations and injury patterns.

Human factors, including age, gender, riding experience, and
risk-taking behavior, have been widely investigated.
\citet{WEBER201447} identified a substantial proportion of
e-bike incidents among riders aged 40--65, while
\citet{HAUSTEIN2016386} reported differences in perceived
safety and incident characteristics across rider groups.
\citet{Fyhri2019Gender} documented gender differences in
e-bike crash risk and increased risk associated with
unfamiliarity with the bicycle. Mobile-phone distraction
has also been associated with road-injury risk among
cyclists and e-bike riders \citep{Ren2021}.
Beyond individual characteristics, injury severity varies
with crash circumstances and traffic conditions.
\citet{Chang2022EbikeSeverity} identified associations
between injury severity and rider, crash, roadway, and
environmental characteristics, while
\citet{Zhu2021PeakTraffic} demonstrated heterogeneity
during peak traffic periods.

Geographic and behavioral contexts further shape e-bike
safety. Rural--urban differences in travel preferences
and e-bike use reflect variations in infrastructure,
trip characteristics, and local conditions
\citep{Hu2021,CherryFishman2021}.
\citet{Jia2026EbikeSeverity} identified age-specific
severity patterns associated with rural locations,
major roads, lighting, and adverse weather.
Safety interventions have consequently emphasized
rider education, helmet use, and regulation.
For example, \citet{Ning2022HelmetCampaign} found
that a helmet-promotion campaign increased overall
helmet use but did not necessarily improve correct
helmet-wearing practices, underscoring the importance
of effective safety education.

Mechanical and electrical components represent another
important dimension of e-bike safety. Battery and
charging-related failures have been associated with
ignition, explosions, and burns \citep{Lei2014},
while tire reliability and puncture-related problems
remain relevant concerns \citep{JaffersonSharma2021}.
\citet{Huertas-Leyva2019} examined e-bike braking
behavior, highlighting the importance of braking
performance given the operating characteristics of
electric bicycles. These studies motivate systematic
investigation of component-related incident mechanisms,
including batteries, drivetrains, brakes, and wheels/tires.

Roadway infrastructure and environmental conditions
also contribute to safety outcomes. Shared-use path
design and operating conditions can influence conflicts
among cyclists, e-bike riders, and pedestrians
\citep{hegde2026completestreet}.
\citet{Chen2010TrafficConflict} associated wider and
smoother paths with fewer traffic conflicts.
Roadway, lighting, and weather conditions have
also been linked to injury-severity differences
\citep{Chang2022EbikeSeverity,Asgarzadeh2018Environment}.
Consistent with these findings, studies of peak-period
e-bike incidents and older riders have identified
heterogeneity associated with traffic conditions,
rural environments, major roads, and adverse weather
\citep{Zhu2021PeakTraffic,Jia2026EbikeSeverity}.

Comparative injury research further demonstrates
differences between e-bikes and other two-wheeled
transportation modes. \citet{Verstappen2021}
identified differences in rider characteristics and
injury patterns between e-bike and conventional
bicycle users, while \citet{QianShi2023} reported
shared and mode-specific injury-severity mechanisms.
Using U.S. injury-surveillance data,
\citet{Burford2024Micromobility} documented injuries
associated with e-bikes, powered scooters,
hoverboards, and conventional bicycles.
These findings highlight the importance of
considering both common and mode-specific
injury characteristics.

Despite these advances, much of the literature relies
on individual structured data sources, potentially
overlooking contextual information embedded in
incident narratives, such as rider behavior,
equipment malfunction, and environmental conditions.
Differences in reporting mechanisms, case composition,
and injury-severity definitions further complicate
comparisons across surveillance systems.
These limitations motivate the present study's
LLM-assisted framework for structuring incident
narratives, harmonizing analytical variables, and
examining source-specific severity associations
alongside supplementary cross-source comparisons.

\section{Methods}

\begin{figure}[!tbp]
    \centering
    \includegraphics[width=0.92\linewidth]{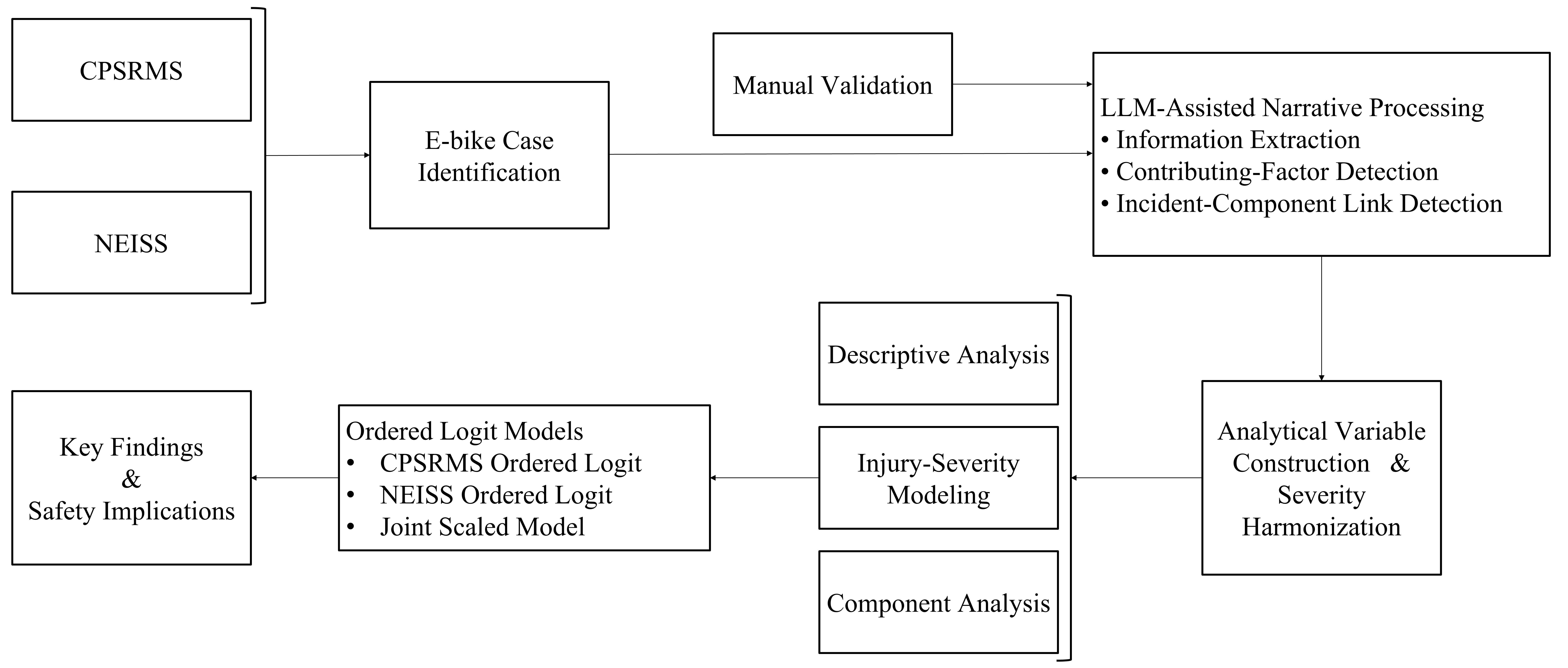}
    \caption{%
    Analytical workflow from e-bike case identification
    and LLM-assisted narrative processing to variable
    construction, descriptive and component analyses,
    injury-severity modeling, and interpretation.}
    \label{fig:study_workflow}
\end{figure}

\begin{figure}[!tbp]
    \centering
    \includegraphics[width=0.78\linewidth]{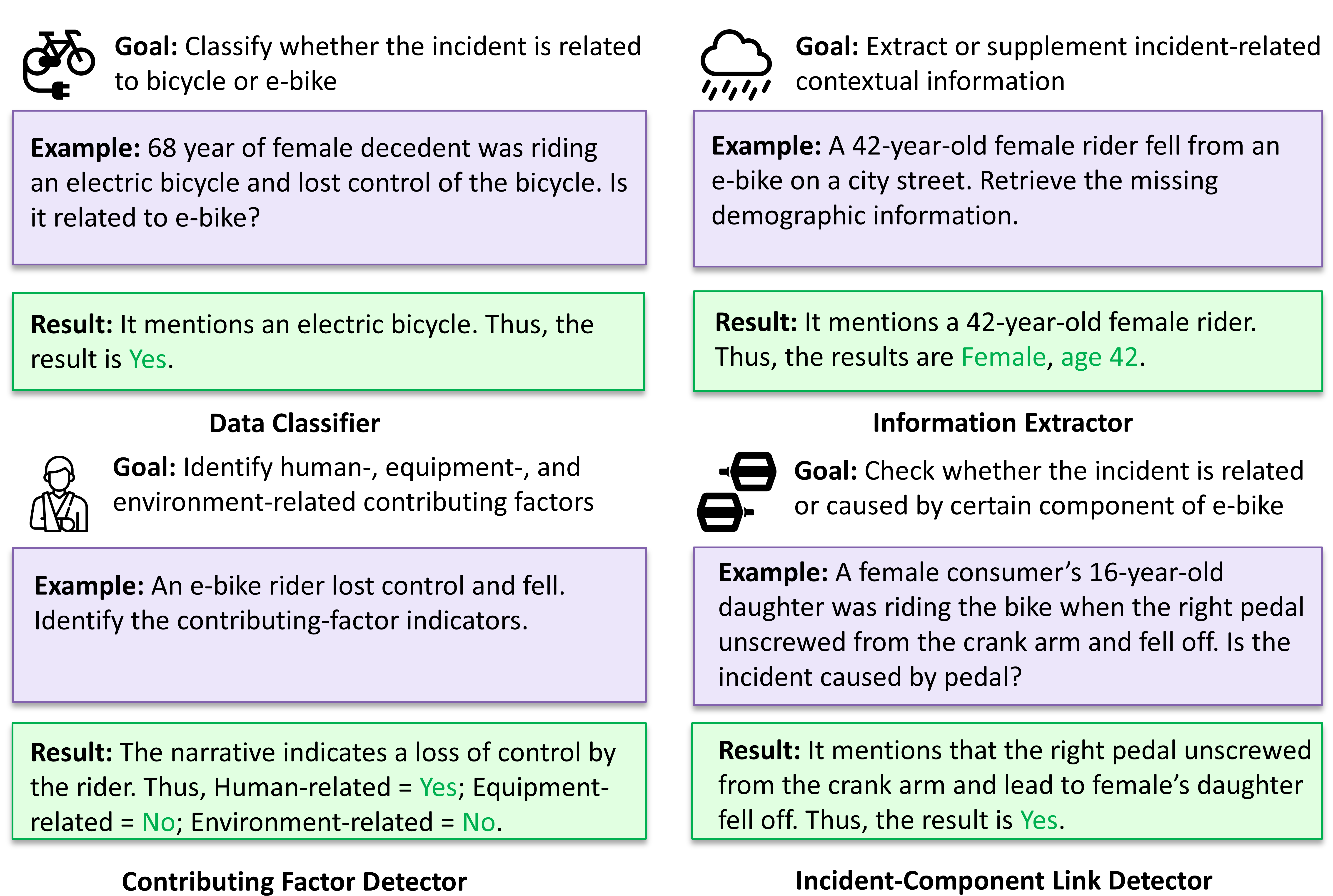}
    \caption{%
    Examples of four LLM agents for e-bike case
    classification, information extraction,
    contributing-factor detection, and
    incident--component link detection.}
    \label{fig:agent_examples}
\end{figure}

Figure~\ref{fig:study_workflow} summarizes the analytical
workflow, from e-bike case identification in CPSRMS and NEISS
to LLM-assisted narrative processing, variable construction,
descriptive and component analyses, and injury-severity modeling.
Four specialized LLM agents perform case classification,
information extraction, contributing-factor detection,
and incident--component link detection, as illustrated in
Figure~\ref{fig:agent_examples}. The resulting structured
variables support source-specific ordered logit models
for CPSRMS and NEISS, along with two supplementary
joint scaled ordered logit specifications.

\subsection{Analyzing Unstructured Incident Narratives with LLMs}

\subsubsection{Data Classifier}

For this study, an electric bicycle (e-bike) is defined as a bicycle equipped with an electric motor that provides pedal assistance or propulsion. The first agent identifies e-bike incidents by interpreting product- and transportation-mode information in the incident narratives, distinguishing them from conventional bicycle incidents. Only records classified as involving e-bikes are included in the analysis, given the limited number of conventional bicycle records identified in both datasets. Because motor specifications and regulatory classifications are not consistently available, the specific e-bike class cannot be verified for each incident.

\subsubsection{Information Extractor}

The second agent extracts or supplements demographic and contextual
information when such information is missing or incomplete in the
structured record. Rider age and gender are taken primarily from structured
fields, with narrative-assisted completion when needed. The agent also
extracts contextual information such as transportation mode, incident
setting, and other narrative details used for data harmonization and
descriptive analysis.
Weather-related variables used in the final ordered logit models are not taken directly from the narrative. In particular, the wet-road indicator is
derived from historical weather information matched to the incident date and geocoded location. Accordingly, the final severity models combine structured demographic, locational, and weather-derived variables with
narrative-derived contributing-factor indicators.

\subsubsection{Contributing-Factor Detector}

Based on the incident narrative, the contributing-factor detector identifies
whether human-related, equipment-related, and environment-related factors
contributed to the incident. These factors are not mutually exclusive.
Each is represented as a separate binary indicator, allowing multiple
contributing factors to be identified within the same case. Human-related indicators capture rider behavior, loss of control,
collision-related behavior, and other human factors. Equipment-related
indicators capture mechanical or electrical issues, including battery,
brake, tire, drivetrain, and other component-related problems.
Environment-related indicators capture roadway, terrain, pavement,
weather, pothole, and other environmental conditions associated with the
incident. These three binary indicators are among the primary narrative-derived
predictors used in the final injury-severity models.

\subsubsection{Incident--Component Link Detector}

Finally, the incident--component link detector evaluates whether an incident is associated with specific e-bike mechanical or
visibility components, as defined in Table~\ref{tab:components_keywords}. For each component, the analysis distinguishes whether the component is related to the incident and whether it is identified as contributing to the incident. This component-level information is used for descriptive analysis and to identify the mechanical
and electrical systems most frequently represented in e-bike incidents.

\begin{table}[!tbp]
\centering
\caption{Mechanical and visibility components explored and corresponding keywords}
\label{tab:components_keywords}

\begingroup
\setstretch{1.0}
\footnotesize
\setlength{\tabcolsep}{3pt}
\renewcommand{\arraystretch}{1.0}

\begin{tabularx}{\linewidth}{@{}
    >{\raggedright\arraybackslash}p{3.7cm}
    >{\raggedright\arraybackslash}X
@{}}
\toprule
\textbf{Mechanical/Visibility Component}
& \textbf{Keyword(s)} \\
\midrule
Brake System & Brake, Pads, Cable \\
Steering System & Handlebar, Stem, Cockpit \\
Pedals & Pedal \\
Drive System & Battery, battery pack, charger/charging system,
motor, controller, chain, belt, sprocket, gear,
derailleur, drivetrain \\
Wheel/Tire & Wheel or Tire, Spokes \\
Front Fork & Front Fork \\
Frame & e-bike Frame \\
Saddles/Seats & Saddle or Seat \\
Visibility & e-bike headlight, tail light, lamp, reflector \\
\bottomrule
\end{tabularx}

\endgroup
\end{table}

\subsection{Incident Severity Analysis: Ordered Logit Models}

We estimate ordered logit models using Biogeme
\citep{bierlaire2020biogeme}, an open-source Python package
for discrete choice modeling. Estimation is performed using a Newton trust-region
algorithm with simple bound constraints. We report robust sandwich
standard errors, log likelihood, the Akaike Information Criterion (AIC),
and the Bayesian Information Criterion (BIC) for the estimated models.

Because CPSRMS and NEISS use different original injury-severity
classifications, the outcome is harmonized into three ordered categories
before model estimation. For CPSRMS, the original seven severity levels
are collapsed into three broader categories. This aggregation is adopted
because several of the original severity levels contain relatively few
observations, which could lead to unstable estimation in a seven-category
ordered logit model. In addition, a three-level outcome provides a more
comparable ordinal structure for cross-source analysis with NEISS. The
CPSRMS categories are defined as follows:

\begin{itemize}[topsep=3pt, partopsep=0pt, itemsep=3pt, parsep=0pt]
  \item \textbf{Minor (1):} original levels 1--2, representing no injury
        or no first aid or medical attention;
  \item \textbf{Outpatient/ED (2):} original levels 3--5, representing
        first aid, evaluation by a medical professional, or
        emergency-department treatment without hospital admission;
  \item \textbf{Serious (3):} original levels 6--7, representing hospital
        admission or death.
\end{itemize}

For NEISS, the treatment-based severity outcome is similarly represented
using three ordered categories: Mild, Moderate, and Severe. The resulting
harmonized outcome therefore ranges from 1, the lowest severity category,
to 3, the highest severity category in both datasets. The harmonization
provides a common ordinal structure while preserving the source-specific
meaning of the underlying severity measures.

For an ordinal response variable $Y$ with ordered categories, the
cumulative ordered logit model is written as
\begin{equation}
  \log\left(
    \frac{P(Y \leq j)}
         {1-P(Y \leq j)}
  \right)
  =
  \tau_j-\mathbf{X}\boldsymbol{\beta},
  \label{eq:ordered_logit}
\end{equation}
where:

\begin{itemize}[topsep=3pt, partopsep=0pt, itemsep=3pt, parsep=0pt]
  \item $P(Y\leq j)$ is the cumulative probability that the outcome is in
        category $j$ or below;
  \item $\tau_j$ denotes the threshold separating adjacent severity
        categories;
  \item $\mathbf{X}$ is the vector of explanatory variables; and
  \item $\boldsymbol{\beta}$ is the corresponding vector of coefficients.
\end{itemize}

Under this parameterization, a positive coefficient is associated with a
greater likelihood of being observed in a higher severity category,
whereas a negative coefficient is associated with a greater likelihood
of being observed in a lower severity category.

Because the harmonized outcome contains three categories, two thresholds
are required. The first threshold, $\tau_1$, is directly estimated. The
second threshold is constructed using an incremental parameter:
\begin{equation}
  \tau_2=\tau_1+\Delta\tau_2,
  \label{eq:threshold_increment}
\end{equation}
where $\Delta\tau_2$ is constrained to preserve the ordering of the
cutpoints.

Four ordered logit specifications are estimated. Models 1 and 2 are
source-specific models for CPSRMS and NEISS, respectively. Their predictor
sets are selected based on model fit, estimation stability, substantive
interpretability, and data availability. These models provide the
source-specific benchmarks and are reported with the two joint models in
Table~\ref{tab:four_model_comparison}.

Models 3 and 4 are joint scaled ordered logit specifications
following the scale-adjustment framework of
\citet{benakiva1990switching}. Both models estimate source-specific
thresholds and a CPSRMS relative-scale parameter, while the NEISS scale
is normalized to one. Model 3 estimates separate source-specific
coefficients except that the age-under-18 and age-18--24 coefficients
are shared as scale anchors. Model 4 is partially unified: the
age-18--24, wet-road, and environment-related coefficients are shared,
whereas age under 18 is estimated separately by source. The remaining
coefficients are allocated to CPSRMS or NEISS according to variable
availability and the source-specific specifications. All four models
are reported in Table~\ref{tab:four_model_comparison}.

In both joint models, the coefficient columns report raw parameters. The CPSRMS linear predictor is multiplied by the relative scale $k$, while the NEISS scale is normalized to one. The reported scale is $k=\exp[\log(k)]$, and its t-statistic tests $H_0:\log(k)=0$, equivalently $H_0: K=1$.

The candidate predictor set was developed by considering demographic,
transportation, locational, temporal, weather-derived, and
narrative-derived variables available across the two datasets.
Alternative specifications were evaluated using the AIC together with
model stability, substantive interpretability, data availability, and
cross-source comparability. Table~\ref{tab:variables_summary_counts}
summarizes the variables constructed and considered during model
development. Because the two surveillance systems differ substantially
in available information and reporting structure, the final retained
predictors are not required to be identical across all source-specific
specifications.

\begin{table}[!tbp]
\centering
\caption{Constructed Variables Considered in the Injury-Severity Modeling Dataset}
\label{tab:variables_summary_counts}

\begingroup
\setstretch{1.0}
\scriptsize
\setlength{\tabcolsep}{2pt}
\renewcommand{\arraystretch}{0.95}

\begin{tabularx}{\textwidth}{@{}
    >{\raggedright\arraybackslash}p{2.15cm}
    >{\raggedright\arraybackslash}X
    >{\raggedright\arraybackslash}p{2.15cm}
    >{\raggedright\arraybackslash}p{3.15cm}
@{}}

\toprule
\textbf{Variable}
& \textbf{Definition and Coding}
& \textbf{Reference Category}
& \textbf{Information Source} \\
\midrule

\multicolumn{4}{@{}l}{\textit{Outcome}} \\

Injury severity
& Harmonized ordered outcome (1--3), with higher
values indicating greater severity. Original CPSRMS
and NEISS classifications are mapped to three levels.
& --
& Source-specific structured severity/treatment fields \\

\addlinespace[2pt]

\multicolumn{4}{@{}l}{\textit{Demographic variables}} \\

Age under 18
& 1 if rider age is below 18; 0 otherwise.
& Age 25--59
& Structured age field; narrative-assisted completion when needed \\

Age 18--24
& 1 if rider age is 18--24; 0 otherwise.
& Age 25--59
& Structured age field; narrative-assisted completion when needed \\

Age 45--59
& 1 if rider age is 45--59; 0 otherwise.
Considered but not retained in the final models.
& Age 25--44 (candidate coding)
& Structured age field; narrative-assisted completion when needed \\

Age above 59
& 1 if rider age is 60 or older; 0 otherwise.
& Age 25--59
& Structured age field; narrative-assisted completion when needed \\

Female
& 1 for female riders; 0 for male riders.
& Male
& Structured demographic field; narrative-assisted completion when needed \\

Black race
& 1 if race is coded as Black; 0 otherwise.
& White
& Structured race field \\

Asian race
& 1 if race is coded as Asian; 0 otherwise.
& White
& Structured race field \\

Other race
& 1 for other reported race categories; 0 otherwise.
& White
& Structured race field \\

\addlinespace[2pt]

\multicolumn{4}{@{}l}{\textit{Transportation variables}} \\

Number of modes
& Number of distinct transportation modes or road
users involved in the incident.
& --
& Structured fields and narrative-derived transportation context \\

\addlinespace[2pt]

\multicolumn{4}{@{}l}{\textit{Location variables}} \\

Street/highway location
& 1 if incident occurred on a street or highway;
0 otherwise.
& Other coded locations
& Structured locale field with harmonized source-specific coding \\

Home location
& 1 if incident occurred at a home, apartment,
condominium, or other residential location;
0 otherwise.
& Other coded locations
& Structured locale field with harmonized source-specific coding \\

Recreation/sport location
& 1 if incident occurred at a recreation or
sport location; 0 otherwise.
& Other coded locations
& Structured locale field with harmonized source-specific coding \\

Rural
& 1 for rural incidents; 0 for urban incidents.
& Urban
& Harmonized geographic classification \\

\addlinespace[2pt]

\multicolumn{4}{@{}l}{\textit{Temporal variables}} \\

Spring
& 1 if incident occurred in spring; 0 otherwise.
& Autumn
& Incident date \\

Summer
& 1 if incident occurred in summer; 0 otherwise.
& Autumn
& Incident date \\

Winter
& 1 if incident occurred in winter; 0 otherwise.
& Autumn
& Incident date \\

Weekend
& 1 if incident occurred on Saturday or Sunday;
0 otherwise.
& Weekday
& Incident date \\

\addlinespace[2pt]

\multicolumn{4}{@{}l}{\textit{Weather-derived variable}} \\

Wet road
& 1 if historical weather-derived road-surface
proxy indicates wet conditions; 0 for dry.
Snow, ice, indoor cases, and unavailable
observations are not coded as wet or dry.
& Dry road
& Historical weather matched to incident date and geocoded location \\

\addlinespace[2pt]

\multicolumn{4}{@{}l}{\textit{Narrative-derived contributing factors}} \\

Human-related factor
& 1 if rider behavior, loss of control,
collision behavior, or another human-related
factor contributed to the incident.
& No human-related factor identified
& Narrative-derived classification \\

Equipment-related factor
& 1 if a mechanical or electrical issue,
including battery, brake, tire, drivetrain,
or other component problems, contributed
to the incident.
& No equipment-related factor identified
& Narrative-derived classification \\

Environment-related factor
& 1 if terrain, pavement, weather,
potholes, or another environmental
factor contributed to the incident.
& No environment-related factor identified
& Narrative-derived classification \\

\bottomrule

\end{tabularx}

\vspace{3pt}

\parbox{\textwidth}{\scriptsize
\textit{Note:} The table summarizes variables considered
during model development. Final predictor sets differ
across specifications. The age 45--59 indicator was
considered but not retained; riders aged 25--59 form
the reference group in the final models.
Models 1 and 2 are source-specific, while Models 3
and 4 are joint scaled specifications
(Table~\ref{tab:four_model_comparison}).
The vehicle-type indicator is excluded because
the final samples contain only e-bike incidents.
}

\endgroup
\end{table}

Observations with missing values in variables required by a given
specification are excluded through complete-case filtering. Missingness
is particularly substantial in CPSRMS for age, gender, street/highway
location, and the weather-derived wet-road indicator, while NEISS shows
notable missingness primarily for street/highway location and wet-road
conditions. Variable-level missingness is summarized in
Table~\ref{tab:missingness}.

\begin{table}[!tbp]
\centering
\caption{Missingness in Key Variables Used in the Injury-Severity Models}
\label{tab:missingness}

\begingroup
\setstretch{1.0}
\footnotesize
\setlength{\tabcolsep}{3pt}
\renewcommand{\arraystretch}{1.0}

\begin{tabular*}{\textwidth}{
@{\extracolsep{\fill}}
l rr rr
}
\toprule
& \multicolumn{2}{c}{\textbf{CPSRMS} ($N=710$)}
& \multicolumn{2}{c}{\textbf{NEISS} ($N=2,376$)} \\

\cmidrule(lr){2-3}
\cmidrule(lr){4-5}

\textbf{Variable}
& \textbf{Missing $N$}
& \textbf{Missing (\%)}
& \textbf{Missing $N$}
& \textbf{Missing (\%)} \\
\midrule

Injury severity
& 0 & 0.0 & 0 & 0.0 \\

Age indicators
& 314 & 44.2 & 0 & 0.0 \\

Female
& 257 & 36.2 & 2 & 0.1 \\

Wet road
& 203 & 28.6 & 584 & 24.6 \\

Street/highway location
& 230 & 32.4 & 814 & 34.3 \\

Human-related factor
& 0 & 0.0 & 0 & 0.0 \\

Equipment-related factor
& 0 & 0.0 & 0 & 0.0 \\

Environment-related factor
& 0 & 0.0 & 0 & 0.0 \\

\bottomrule
\end{tabular*}

\vspace{2pt}

\parbox{\textwidth}{\footnotesize
\textit{Note:} Missingness is calculated using the full
classified samples (CPSRMS: $N=710$; NEISS: $N=2,376$).
Age indicators share the same underlying missingness pattern.
A zero value for a narrative-derived factor indicates that
the factor was not identified and is not treated as missing.
Complete-case sample sizes vary by model specification:
278 for Model 1, 1,181 for Model 2, and 1,459 for Models 3 and 4.
}

\endgroup
\end{table}

The descriptive, cause, and component analyses use the full classified
samples of 710 CPSRMS and 2,376 NEISS e-bike incidents. After
specification-specific complete-case filtering, Model 1 uses 278 CPSRMS
observations and Model 2 uses 1,181 NEISS observations. Models 3 and 4
use the stacked sample of 1,459 observations. These sample sizes are
reported in Table~\ref{tab:four_model_comparison}.

Because the retained predictors differ across specifications, variables
that are useful within one surveillance system are not necessarily
retained in the other. Female, rural, and street/highway indicators are
retained for NEISS, whereas spring, weekend, human-related,
equipment-related, and recreation/sport indicators are retained for
CPSRMS. Wet-road and environment-related factors are included for both
sources. These differences reflect data availability, reporting
structure, substantive relevance, and empirical fit.

Finally, the joint models are used as supplementary assessments of cross-source comparability rather than as the primary models for substantive interpretation. Likelihood-ratio tests support selected coefficient-sharing restrictions but reject joint equality of the two source thresholds. The source-specific models therefore remain the primary modeling approach, while Model 4 provides a parsimonious supplementary joint specification.

\section{Experiments}

The experiments in this section are designed to address the following
research questions:

\begin{enumerate}
    \item What demographic and contextual characteristics are observed
    among e-bike-related incidents in CPSRMS and NEISS?

    \item How frequently are human-related, equipment-related, and
environment-related contributing factors identified in e-bike incidents?

    \item What age- and gender-related patterns emerge in the most
    frequently reported causes of e-bike incidents?

    \item What injury causes are most frequently associated with human-,
    equipment-, and environment-related e-bike incident descriptions?

    \item How is injury severity distributed across age and gender groups
    within CPSRMS and NEISS?

    \item Which demographic, locational, weather-derived, and
    narrative-derived factors are associated with e-bike injury severity
    within CPSRMS and NEISS?

    \item How do the source-specific severity relationships differ between CPSRMS and NEISS, and what additional evidence is provided by the joint scaled ordered logit models?

    \item Which mechanical and visibility components are most frequently
    associated with e-bike incidents, and how accurately can the
    LLM-based incident--component link detector identify these
    associations?
\end{enumerate}

\subsection{Experiment Settings}
\subsubsection{Datasets}

This study uses two complementary CPSC injury-surveillance
datasets, CPSRMS and NEISS, covering 2017--2023.

CPSRMS contains consumer-product incident reports submitted
by consumers, healthcare providers, manufacturers, government
agencies, and other reporting entities. Records include structured
information on incident date, location, severity, and rider
demographics, together with detailed free-text narratives
describing incident circumstances, rider behavior, and equipment
problems. An LLM-based classification agent identified 710
e-bike-related incidents for analysis.

NEISS provides standardized emergency-department records,
including product codes, diagnoses, injury locations,
demographic characteristics, and treatment outcomes,
supplemented by short narratives (\textit{narr}).
Its injury data are collected from approximately 100
hospitals selected as a probability sample of more than
5,000 U.S. hospitals with emergency departments.
For 2017--2023, we identified 2,376 e-bike-related cases.

Unlike conventional transportation databases such as FARS
\citep{NHTSA_FARS} and CRSS \citep{NHTSA_CRSS},
CPSRMS contains relatively detailed unstructured narratives,
while NEISS emphasizes standardized clinical records.
Their integration supports complementary analyses of
injury causes, rider characteristics, component involvement,
and severity outcomes. The full classified samples were
used for descriptive, cause, and component analyses,
whereas ordered logit models used specification-specific
complete-case subsamples, as reported in
Table~\ref{tab:four_model_comparison}.

\subsubsection{LLM Model Details}

In this study, we apply the GPT-4 \citep{openai2024gpt4technicalreport}
model to execute the four agents described in the LLM-based
narrative-analysis framework. We use zero-shot prompting. Representative prompt examples and corresponding expected
outputs are provided in Table~\ref{tab:prompt_examples}.

\begin{table}[!tbp]
\centering
\caption{Examples of LLM Prompts and Expected Outputs}
\label{tab:prompt_examples}

\begingroup
\setstretch{1.0}
\footnotesize
\setlength{\tabcolsep}{3pt}
\renewcommand{\arraystretch}{1.0}
\renewcommand{\tabularxcolumn}[1]{m{#1}}

\begin{tabularx}{\textwidth}{
|>{\centering\arraybackslash}m{2.3cm}
|>{\raggedright\arraybackslash}X
|>{\centering\arraybackslash}m{2.75cm}|}

\hline
\textbf{Task}
& \centering\textbf{Example Prompt}
& \textbf{Expected Output}
\tabularnewline
\hline

Transportation Mode
&
Please retrieve the transportation mode in the incident:
``A 66-year-old male riding an electric bicycle eastbound on a road
was seriously injured after colliding with a southbound vehicle
at the intersection.''
&
\shortstack[c]{electric \\ bicycle}
\\
\hline

Demographic Information
&
Please retrieve the demographic information in the incident:
``42YOF PT RIDING AN E BIKE, LOST CONTROL \& CRASHED BIKE.
DX: HEAD CONTUSION, CHI.''
&
\shortstack[c]{Female \\ Age 42}
\\
\hline

Contributing Factors
&
Determine whether human-related, equipment-related, and
environment-related factors contributed to the incident:
``9 yom riding an electric bike down a hill, lost control and
hit a car. dx orbital fx.''
&
\shortstack[c]{
Human = Yes \\
Equipment = No \\
Environment = No
}
\\
\hline

Incident--Component Link
&
Identify the e-bike component related to the incident:
``PEDAL BROKE''
&
Pedals = Yes
\\
\hline

\end{tabularx}
\endgroup
\end{table}

The agents extracted or supplemented transportation context, demographic
information, incident characteristics, and narrative-derived contributing
factors from the incident narratives. In particular, the contributing-factor
detector generated separate human-, equipment-, and environment-related
binary indicators. Weather-derived variables used in the injury-severity
models were constructed separately rather than extracted from the narratives;
specifically, the wet-road indicator was derived from historical weather
information matched to the incident date and geocoded location.

These candidate variables were initially considered because demographic,
transportation, environmental, and incident-related characteristics are
frequently examined in the injury-severity literature
\citep{WEBER201447, Chen2010, Ma2023}. Final predictor sets were selected
based on model fit, estimation stability, substantive interpretability,
data availability, and the reporting characteristics of each source.
Models 1 and 2 provide the CPSRMS- and NEISS-specific specifications,
whereas Models 3 and 4 provide the joint scaled specifications. All four
models are reported in Table~\ref{tab:four_model_comparison}.

\subsubsection{Evaluation Metrics}

To evaluate the performance of the incident--component link detector,
we constructed a manually reviewed validation sample of 300 e-bike
incidents, comprising 150 CPSRMS and 150 NEISS cases. Within each
data source, the sample was enriched with component-positive cases
to ensure adequate representation of relatively infrequent component
classes while retaining predicted-negative cases for false-negative
assessment. The evaluation framework and classification results
are summarized in Table~\ref{tab:confusion_matrix}.

\begin{table}[!tbp]
\centering

\caption{Confusion Matrix and Component Classification
Performance for CPSRMS and NEISS}

\label{tab:confusion_matrix}
\label{tab:cpsrms_f1_results}
\label{tab:neiss_f1_results}

\begingroup
\setstretch{1.0}
\footnotesize
\setlength{\tabcolsep}{3pt}
\renewcommand{\arraystretch}{1.0}

\noindent
\textbf{Panel A. Confusion Matrix}

\vspace{3pt}

\begin{tabular}{lcc}
\toprule
& \textbf{Predicted Positive}
& \textbf{Predicted Negative} \\
\midrule

\textbf{Actual Positive}
& True Positive (TP)
& False Negative (FN) \\

\textbf{Actual Negative}
& False Positive (FP)
& True Negative (TN) \\

\bottomrule
\end{tabular}

\vspace{10pt}

\noindent
\textbf{Panel B. CPSRMS Classification Performance}

\vspace{3pt}

\begin{tabularx}{\textwidth}{@{}
    >{\raggedright\arraybackslash}p{3.0cm}
    *{6}{>{\centering\arraybackslash}X}
@{}}

\toprule
\textbf{Component Class}
& \textbf{TP}
& \textbf{FP}
& \textbf{FN}
& \textbf{Precision}
& \textbf{Recall}
& \textbf{F1 Score} \\
\midrule

Brake System
& 13 & 3 & 3 & 0.81 & 0.81 & 0.81 \\

Pedals
& 36 & 0 & 1 & 1.00 & 0.97 & 0.99 \\

Wheel/Tire
& 20 & 4 & 1 & 0.83 & 0.95 & 0.89 \\

Saddle/Seat
& 3 & 3 & 0 & 0.50 & 1.00 & 0.67 \\

Front Fork
& 6 & 5 & 1 & 0.55 & 0.86 & 0.67 \\

Visibility
& 3 & 0 & 0 & 1.00 & 1.00 & 1.00 \\

Drive System
& 34 & 1 & 2 & 0.97 & 0.94 & 0.96 \\

Frame
& 8 & 1 & 1 & 0.89 & 0.89 & 0.89 \\

Steering System
& 5 & 1 & 1 & 0.83 & 0.83 & 0.83 \\

\midrule

\textbf{Weighted F1 Score}
& \multicolumn{6}{c}{\textbf{0.91}} \\

\bottomrule
\end{tabularx}

\vspace{10pt}

\noindent
\textbf{Panel C. NEISS Classification Performance}

\vspace{3pt}

\begin{tabularx}{\textwidth}{@{}
    >{\raggedright\arraybackslash}p{3.0cm}
    *{6}{>{\centering\arraybackslash}X}
@{}}

\toprule
\textbf{Component Class}
& \textbf{TP}
& \textbf{FP}
& \textbf{FN}
& \textbf{Precision}
& \textbf{Recall}
& \textbf{F1 Score} \\
\midrule

Brake System
& 12 & 2 & 0 & 0.86 & 1.00 & 0.92 \\

Pedals
& 5 & 0 & 0 & 1.00 & 1.00 & 1.00 \\

Wheel/Tire
& 22 & 1 & 2 & 0.96 & 0.92 & 0.94 \\

Saddle/Seat
& 1 & 1 & 0 & 0.50 & 1.00 & 0.67 \\

Front Fork
& 1 & 0 & 0 & 1.00 & 1.00 & 1.00 \\

Visibility
& 0 & 0 & 0 & -- & -- & -- \\

Drive System
& 8 & 1 & 0 & 0.89 & 1.00 & 0.94 \\

Frame
& 2 & 0 & 2 & 1.00 & 0.50 & 0.67 \\

Steering System
& 4 & 1 & 6 & 0.80 & 0.40 & 0.53 \\

\midrule

\textbf{Weighted F1 Score}
& \multicolumn{6}{c}{\textbf{0.86}} \\

\bottomrule
\end{tabularx}

\vspace{4pt}

\parbox{\textwidth}{\footnotesize
\textit{Note:} TP = True Positive;
FP = False Positive;
FN = False Negative;
TN = True Negative.
Precision, recall, and F1 scores are reported
for each mechanical or visibility component class.
Weighted F1 scores are 0.91 for CPSRMS
and 0.86 for NEISS.
}

\endgroup
\end{table}

Each selected incident was manually reviewed for all nine component
classes listed in Table~\ref{tab:components_keywords}. A component
was labeled as positive when it was explicitly mentioned, clearly
involved, or meaningfully connected to the incident or injury
process; causal contribution was not required for this
component-link classification task. The manually coded labels
served as the reference standard for evaluating agent predictions.

As illustrated in Panel A of Table~\ref{tab:confusion_matrix},
the confusion matrix distinguishes four possible outcomes:

\begin{itemize}
    \item \textbf{TP (True Positive):} The component was correctly
    identified as related to the incident.

    \item \textbf{FP (False Positive):} The component was predicted
    as related but was not supported by manual review.

    \item \textbf{TN (True Negative):} The component was correctly
    identified as not related to the incident.

    \item \textbf{FN (False Negative):} The component was identified
    as related during manual review but was missed by the agent.
\end{itemize}

Separate confusion matrices were constructed for each of the nine
component classes in CPSRMS and NEISS, yielding 18 class-specific
evaluations. Classification performance was assessed using
precision, recall, and the F1 score, defined as
\begin{equation}
\text{Precision} = \frac{TP}{TP + FP},
\qquad
\text{Recall} = \frac{TP}{TP + FN}.
\end{equation}
\begin{equation}
F1 =
\frac{2 \times \text{Precision} \times \text{Recall}}
{\text{Precision} + \text{Recall}}.
\end{equation}

To summarize overall performance across the $N=9$ component
classes, we calculated the support-weighted F1 score:
\begin{equation}
F1_{\mathrm{weighted}} =
\frac{\sum_{i=1}^{N} F1_i \times \mathrm{Support}_i}
{\sum_{i=1}^{N} \mathrm{Support}_i}.
\end{equation}

Here, $F1_i$ denotes the F1 score for component class $i$,
and $\mathrm{Support}_i$ is the number of manually identified
positive instances for that class. Classes with no positive
reference instances in a given dataset do not contribute
to the weighted F1 score.

\subsection{Classification Results}

Panels B and C of Table~\ref{tab:confusion_matrix} present the
component-level classification results for CPSRMS and NEISS,
respectively.

The incident--component link detector achieved weighted F1
scores of 0.91 for CPSRMS and 0.86 for NEISS, indicating strong
overall agreement with the manually reviewed reference labels.
Performance varied across component classes, with particularly
high F1 scores for pedals, drive systems, and wheel/tire-related
cases. Lower performance for some less frequently represented
classes suggests that component-specific ambiguity and sparse
positive examples remain important challenges.

Overall, these results demonstrate the potential of LLM-assisted
extraction to transform heterogeneous transportation-injury
narratives into structured component-level information.

\subsection{Visualization}
\subsubsection{Initial Exploration}

As described in the LLM-based narrative analysis framework, incidents
were classified for subsequent analysis. The final classified e-bike
samples contained 710 CPSRMS incidents and 2,376 NEISS incidents.
These full classified samples were used for the descriptive analyses
presented in this section.

\begin{figure}[!tbp]
    \centering
    \includegraphics[width=\linewidth]{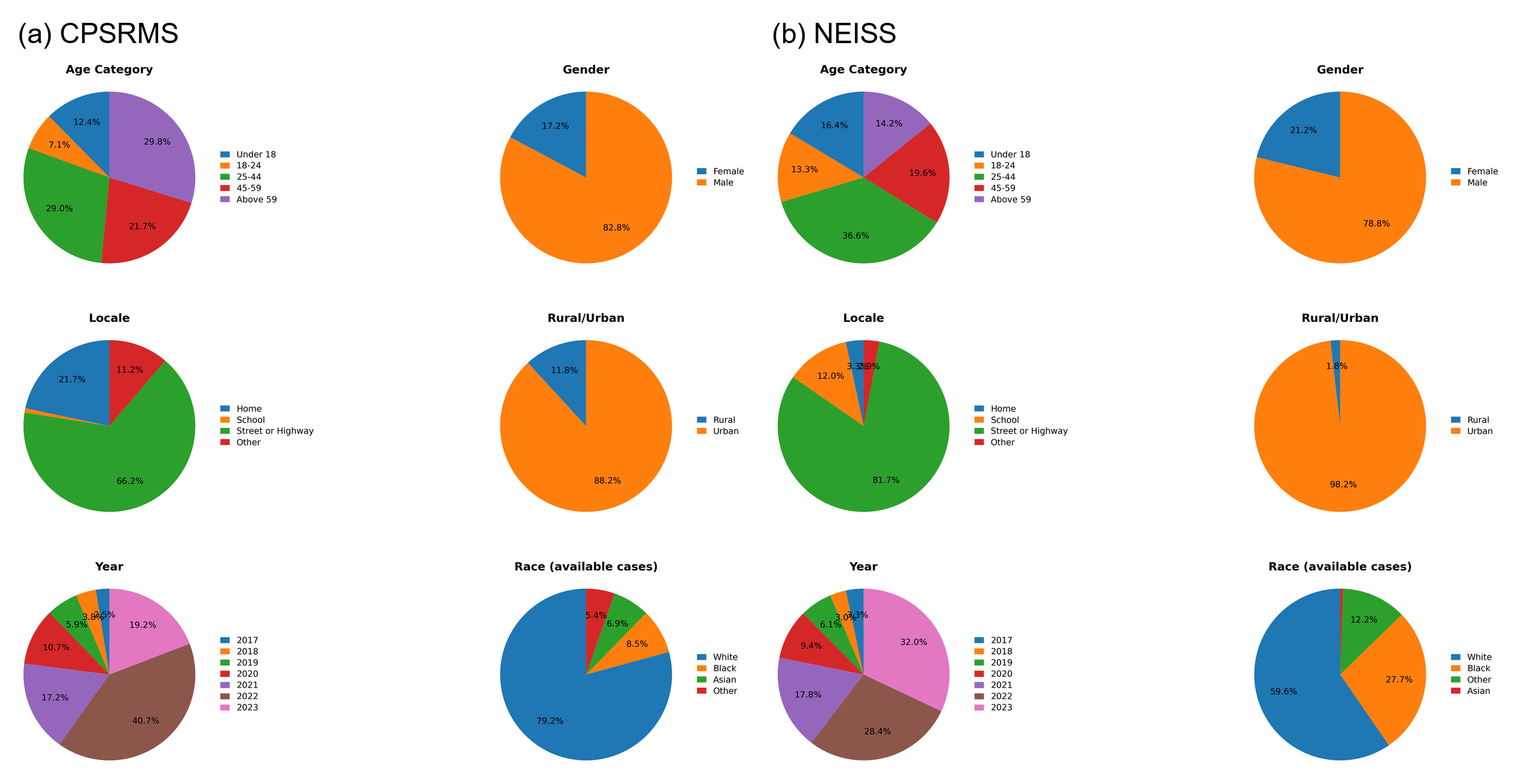}
    \caption{%
    Distributions of age, gender, incident locale,
    rural/urban classification, year, and race
    in (a) CPSRMS and (b) NEISS.}
    \label{fig::distribution_combined}
\end{figure}

The visualizations summarize demographic, temporal, and contextual
patterns in the two e-bike datasets. In CPSRMS, most incidents occurred
on streets or highways and in urban areas. Incidents were most frequently
recorded in 2022, while riders aged 25--44 and those above age 59
represented relatively large shares of the sample. Among CPSRMS cases
with available gender information, male riders accounted for the majority
of e-bike incidents. In NEISS, street/highway and urban incidents were
also predominant. Riders aged 25--44 constituted the largest age group,
and male riders represented the majority of cases. White patients
accounted for the largest reported racial group among cases with
available race information. The number of recorded NEISS e-bike
incidents increased substantially in the later years of the study
period, with the highest frequency observed in 2023.
Figure~\ref{fig::distribution_combined} summarizes these descriptive
patterns for CPSRMS and NEISS.

\subsubsection{Causes of E-Bike Incidents}

\begin{figure}[!tbp]
    \centering
    \includegraphics[width=\linewidth]{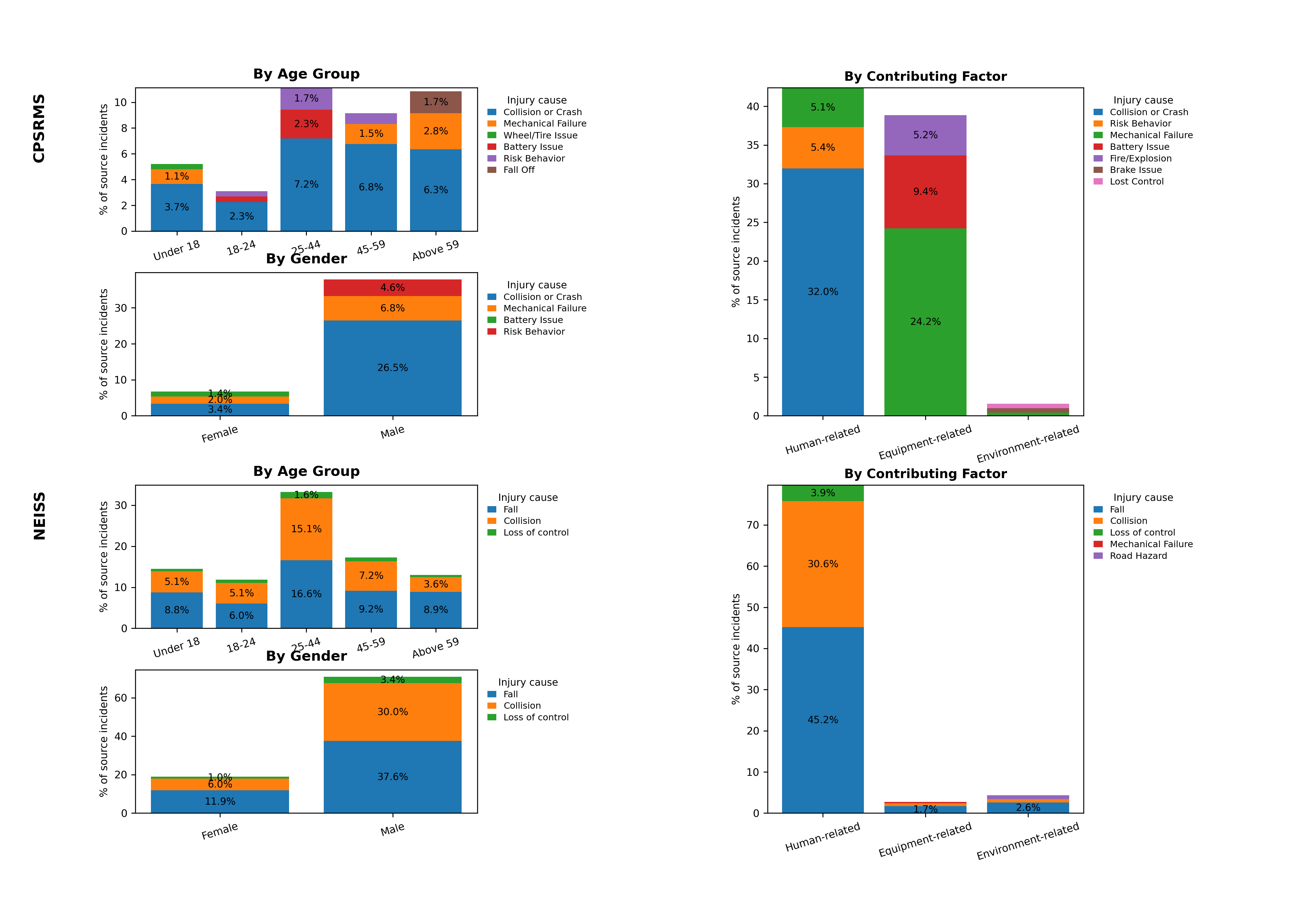}
    \caption{%
    Frequently identified injury causes in CPSRMS
    (upper panels) and NEISS (lower panels), grouped
    by age and gender (left) and contributing-factor
    categories (right). Human-, equipment-, and
    environment-related indicators are not
    mutually exclusive.}
    \label{fig::causes_age_gender_combined}
\end{figure}

Figure~\ref{fig::causes_age_gender_combined} presents the distribution
of frequently identified injury causes across age and gender groups
(left panels) and contributing-factor categories (right panels) for
CPSRMS and NEISS.

The two data sources exhibit different cause profiles. In CPSRMS,
collision- or crash-related events are frequently identified across
demographic groups, together with equipment-related causes such as
mechanical failures, battery issues, wheel/tire problems, and other
component malfunctions. In NEISS, falls and collisions account for
a substantially larger share of observed cases, while loss of
control also appears repeatedly across age and gender groups.
These differences are consistent with the contrasting reporting
structures of the two datasets: CPSRMS contains more detailed
product- and equipment-related incident narratives, whereas
NEISS primarily records emergency-department injury events.

The right panels of Figure~\ref{fig::causes_age_gender_combined}
further summarize injury causes according to human-related,
equipment-related, and environment-related contributing factors.
In CPSRMS, human-related incidents are dominated by collision-
and crash-related descriptions, whereas equipment-related incidents
include mechanical failures, battery problems, fire or explosion
events, brake issues, wheel/tire problems, and other component-related
failures. Environment-related factors are less frequently identified
and include roadway, surface, and weather conditions.

In NEISS, human-related incidents are dominated by falls and
collisions, whereas equipment-related incidents account for a
smaller share of the sample. Environment-related factors are
also less frequent and capture roadway, surface, or other
environmental conditions associated with the incident.
The underlying human-, equipment-, and environment-related
indicators are coded independently and may co-occur within
individual incidents.

\subsubsection{Injury Severity Analysis}

\begin{figure}[!tbp]
    \centering
    \includegraphics[width=0.92\linewidth]{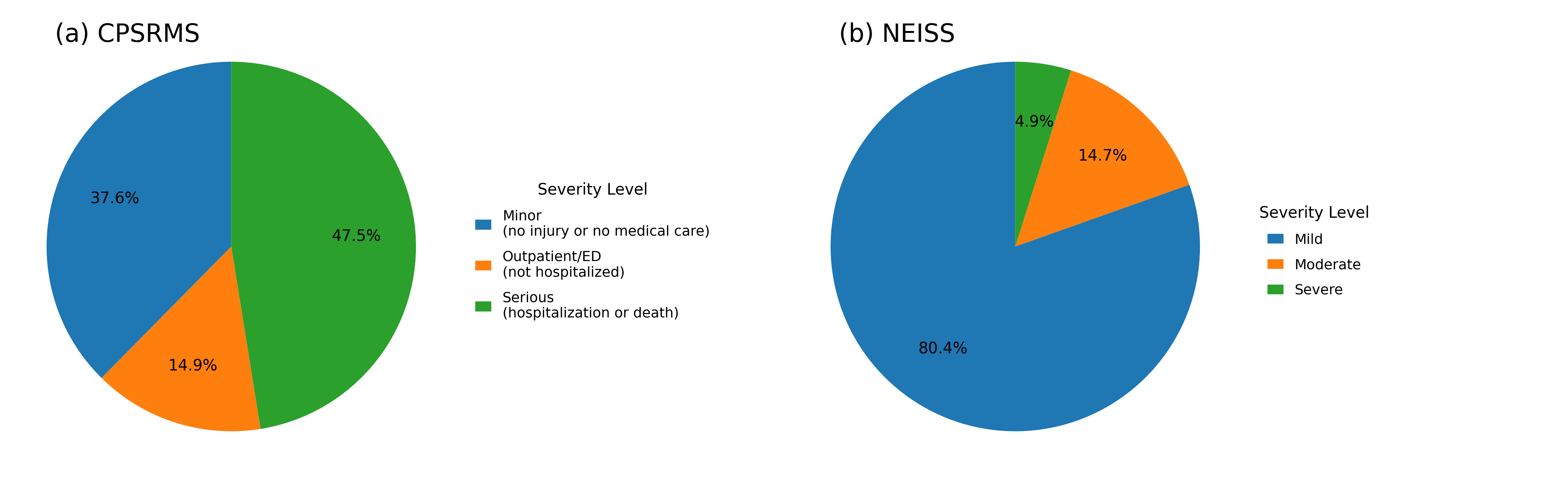}
    \caption{%
    Injury-severity distributions in (a) CPSRMS
    (Minor, Outpatient/ED, Serious) and (b) NEISS
    (Mild, Moderate, Severe). Severity categories
    are source-specific and not directly equivalent.}
    \label{fig:severitydistribution_combined}
\end{figure}

We first examine the descriptive distribution of injury severity in
the two e-bike datasets. Because CPSRMS and NEISS originate from
different surveillance systems, the severity labels retain
source-specific terminology even though both are represented
using three ordered levels. For CPSRMS, the three categories are
Minor, Outpatient/ED, and Serious, whereas NEISS uses Mild,
Moderate, and Severe.
Figure~\ref{fig:severitydistribution_combined} summarizes these
distributions for both datasets.

The severity distributions should be interpreted within the reporting
context of their respective data sources rather than compared as
directly equivalent clinical severity measures. CPSRMS contains a
substantially larger share of incidents classified in the Serious
category, whereas NEISS observations are strongly concentrated in
the Mild treatment category. This contrast reflects differences in
reporting mechanisms, case composition, and severity coding between
consumer-product incident reports and emergency-department
surveillance records.

\begin{figure}[!tbp]
    \centering
    \includegraphics[width=\linewidth]{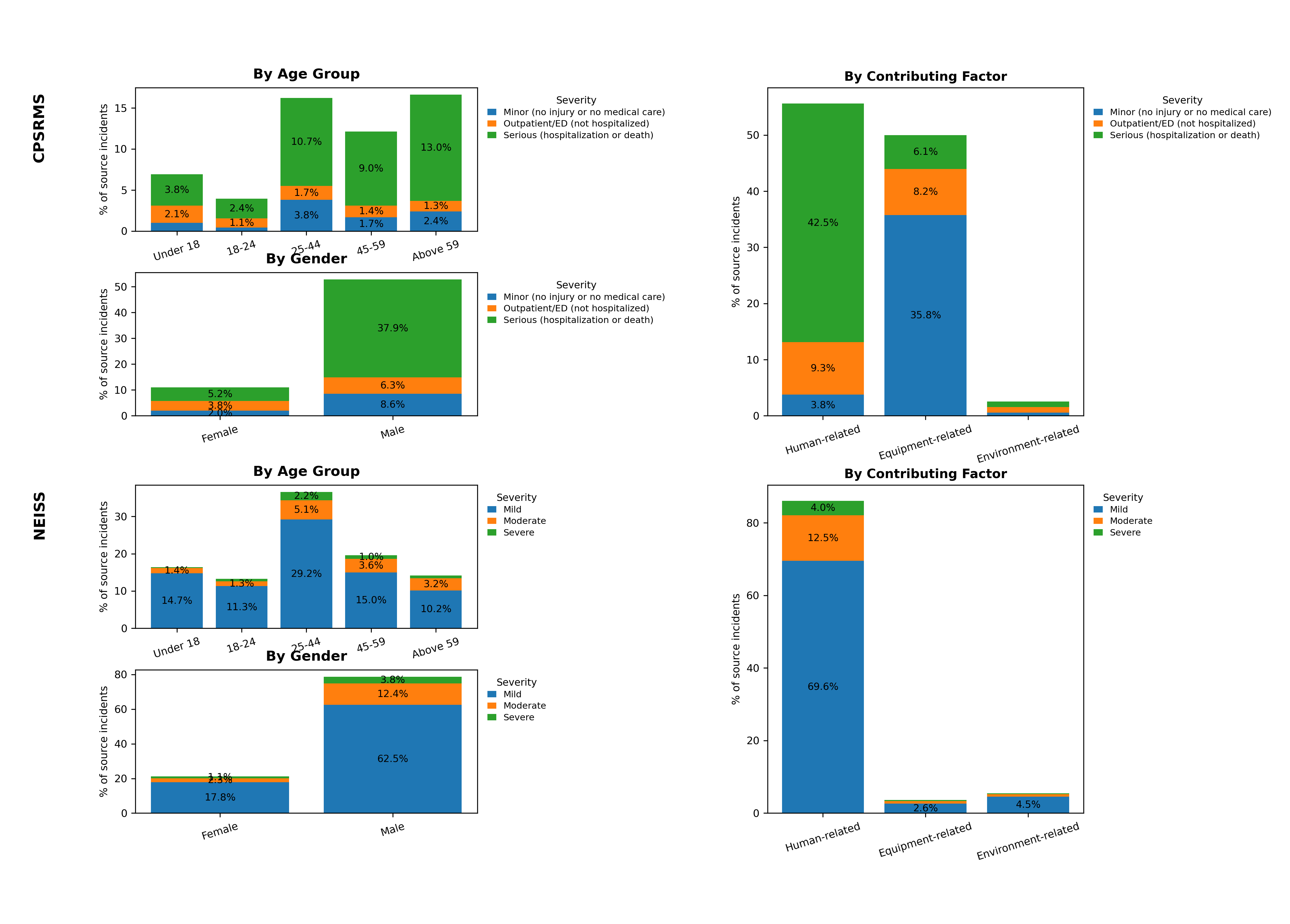}
    \caption{%
    Injury severity by age and gender (left panels)
    and contributing factors (right panels) in
    CPSRMS (upper) and NEISS (lower).
    Severity categories are source-specific, and
    contributing-factor indicators are not
    mutually exclusive.}
    \label{fig:severity_age_gender_combined}
\end{figure}

Figure~\ref{fig:severity_age_gender_combined} further summarizes
injury-severity patterns across age groups and gender (left panels)
and contributing-factor categories (right panels) for CPSRMS and
NEISS.

In CPSRMS, Serious cases account for a relatively large share of
incidents across several age groups, particularly among older
riders. Among CPSRMS cases with known gender, male riders account
for a larger share than female riders in each severity category.
In NEISS, the Mild category dominates across all age and gender
groups, while Moderate and Severe outcomes constitute much
smaller shares of the sample.

These descriptive patterns motivate the subsequent multivariate
analysis, which evaluates whether age, gender, location,
weather-derived conditions, and narrative-derived incident
factors remain associated with injury severity after controlling
for the other covariates.

The right panels of Figure~\ref{fig:severity_age_gender_combined}
examine the descriptive relationship between contributing-factor
indicators and injury severity. Specifically, they summarize
severity among incidents identified as human-related,
equipment-related, and environment-related. Because these
three indicators are coded independently, an individual
incident may appear in more than one contributing-factor group.

In CPSRMS, human-related incidents contain a comparatively
large share of Serious outcomes, whereas equipment-related
incidents are more frequently represented in the Minor category.
Environment-related incidents account for a smaller share of
the full classified sample. In NEISS, the Mild category dominates
across human-, equipment-, and environment-related incidents,
while Moderate and Severe outcomes represent substantially
smaller shares.

The contributing-factor groups shown in the right panels of
Figure~\ref{fig:severity_age_gender_combined} are not mutually
exclusive, as multiple factors may be identified within the
same incident. These patterns provide descriptive evidence
only and do not establish causal relationships.

\subsubsection{Link between Incident and E-Bike Components}

Compared with conventional bicycles, e-bikes contain more complex
mechanical and electrical systems, including batteries, charging
systems, motors, drivetrain components, braking systems, and other
mechanical hardware. It is therefore important to distinguish
between components that are merely associated with an incident
and components that are identified as directly contributing to
the incident. Such information can support component design,
maintenance practices, product-safety monitoring, and the
development of targeted safety standards.

\begin{figure}[!tbp]
    \centering
    \includegraphics[width=0.95\linewidth]{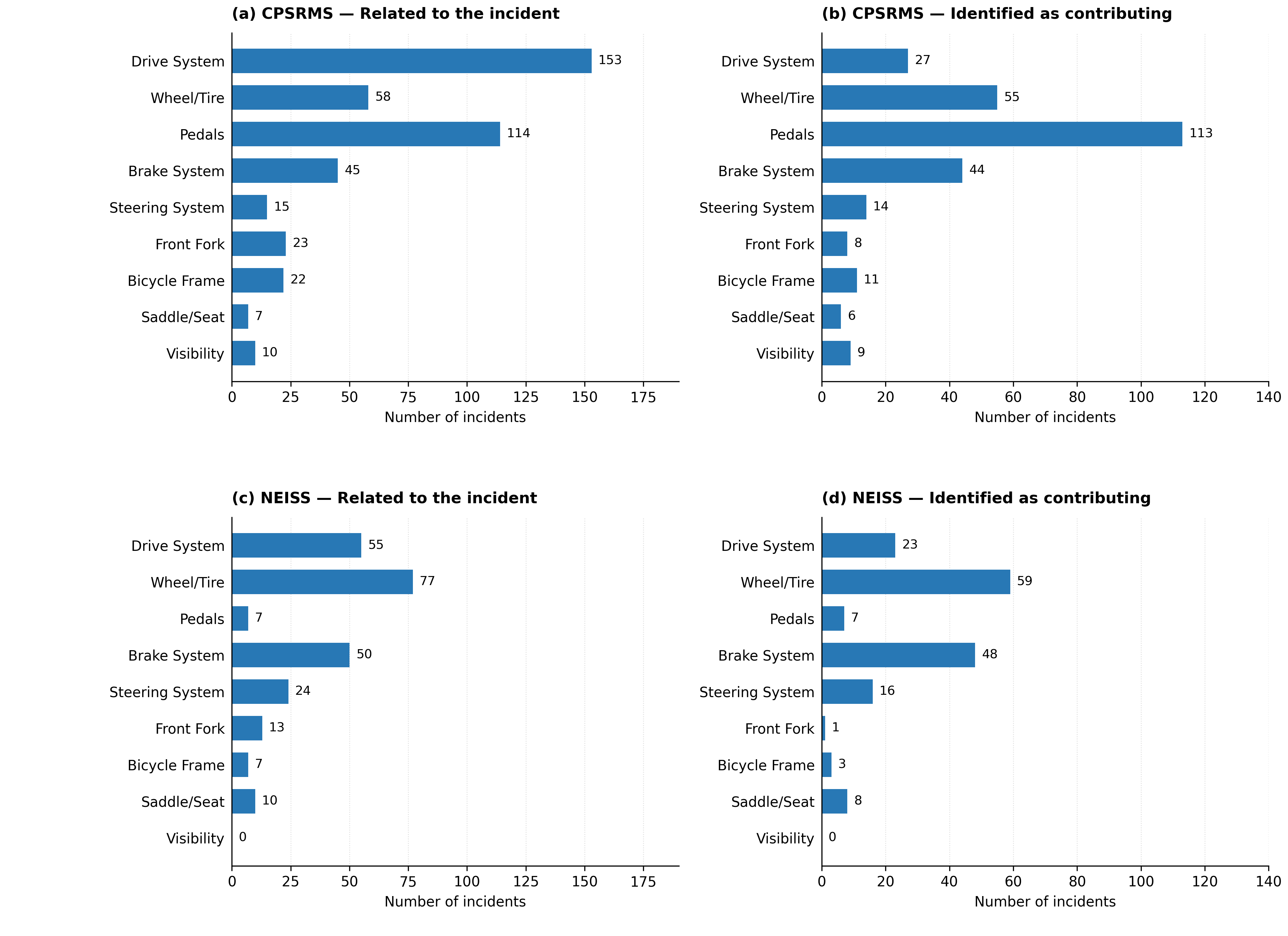}
    \caption{%
    Mechanical and visibility components related
    to incidents (left) and identified as
    contributing (right) in CPSRMS (upper panels)
    and NEISS (lower panels).}
    \label{fig::components_combined}
\end{figure}

Using the incident narratives, we identify links between e-bike
incidents and nine mechanical or visibility component classes.
For each component, two descriptive indicators are considered:
whether the component is related to or involved in the incident,
and whether the component is identified as contributing to
the incident. Figure~\ref{fig::components_combined} summarizes
these patterns for CPSRMS (upper panels) and NEISS (lower panels).
The left panels present components related to incidents, whereas
the right panels present components identified as contributing
to incidents.

The two data sources exhibit distinct component profiles. In CPSRMS,
the drive system is the most frequently associated component, followed
by pedals, wheels/tires, and brake systems. The high frequency of
drive-system involvement reflects the broader range of electrical and
propulsion-related issues captured in the CPSRMS narratives, including
battery, charging, motor, and drivetrain problems. When attention is
restricted to components identified as contributing to incidents,
pedals are the most frequently represented component, followed by
wheels/tires, brake systems, and the drive system.

In NEISS, wheel/tire involvement is most frequently identified,
followed by the drive system and brake system. Wheels/tires and
brake systems are also the most frequently identified contributing
components. Steering, front-fork, frame, and saddle/seat involvement
occurs less frequently, while no NEISS cases in the classified sample
were identified as involving bicycle-specific visibility equipment.

These descriptive differences are consistent with the contrasting
reporting structures of the two datasets. CPSRMS contains more
detailed product- and equipment-focused narratives and therefore
captures a wider range of electrical and mechanical problems.
NEISS narratives are shorter and clinically oriented, resulting
in greater emphasis on crash mechanisms and immediately
observable mechanical involvement.

\subsection{Predicting E-Bike Incident Severity}

We examine associations between injury severity and demographic,
locational, temporal, weather-derived, and narrative-derived factors
using four ordered logit specifications (Table~\ref{tab:four_model_comparison}).
Models 1 and 2 are source-specific models for CPSRMS and NEISS,
respectively, while Models 3 and 4 are supplementary joint scaled
specifications assessing cross-source comparability. Positive
coefficients indicate associations with higher severity categories,
whereas negative coefficients indicate associations with lower
severity categories.

\begin{table}[!tbp]
\centering
\caption{Estimation Results for the Source-Specific and Joint Scaled Ordered Logit Models}
\label{tab:four_model_comparison}

\begingroup
\setstretch{1.0}
\scriptsize
\setlength{\tabcolsep}{2.6pt}
\renewcommand{\arraystretch}{1.0}

\begin{tabular*}{\textwidth}{@{\extracolsep{\fill}}l cc cc cc cc}
\toprule
& \multicolumn{2}{c}{\textbf{Model 1}}
& \multicolumn{2}{c}{\textbf{Model 2}}
& \multicolumn{2}{c}{\textbf{Model 3}}
& \multicolumn{2}{c}{\textbf{Model 4}} \\

& \multicolumn{2}{c}{CPSRMS only}
& \multicolumn{2}{c}{NEISS only}
& \multicolumn{2}{c}{\shortstack{Joint: separate $\beta$\\+ relative scale}}
& \multicolumn{2}{c}{\shortstack{Joint: source allocated/\\partially unified + relative scale}} \\

\cmidrule(lr){2-3}
\cmidrule(lr){4-5}
\cmidrule(lr){6-7}
\cmidrule(lr){8-9}

\textbf{Model information}
& \textbf{Coeff.} & \textbf{t-stat}
& \textbf{Coeff.} & \textbf{t-stat}
& \textbf{Coeff.} & \textbf{t-stat}
& \textbf{Coeff.} & \textbf{t-stat} \\
\midrule

Female
& &
& -0.403** & -2.05
& \shortstack{N: -0.404**}
& \shortstack{N: -2.06}
& \shortstack{N: -0.402**}
& \shortstack{N: -2.05} \\

Rural
& &
& 0.325 & 0.85
& \shortstack{N: 0.326}
& \shortstack{N: 0.85}
& \shortstack{N: 0.330}
& \shortstack{N: 0.86} \\

Spring
& -0.978** & -2.13
& &
& \shortstack{C: -0.516}
& \shortstack{C: -1.49}
& \shortstack{C: -0.276}
& \shortstack{C: -1.34} \\

Weekend
& 0.552 & 1.23
& &
& \shortstack{C: 0.294}
& \shortstack{C: 1.18}
& \shortstack{C: 0.161}
& \shortstack{C: 1.03} \\

Age 18--24
& -0.715 & -1.06
& -0.282 & -1.23
& \shortstack{C: -0.314\\N: -0.314}
& \shortstack{C: -1.58\\N: -1.58}
& \shortstack{C: -0.247\\N: -0.247}
& \shortstack{C: -1.41\\N: -1.41} \\

Age above 59
& 0.227 & 0.52
& 0.395** & 2.14
& \shortstack{C: 0.130\\N: 0.392**}
& \shortstack{C: 0.51\\N: 2.13}
& \shortstack{N: 0.400**}
& \shortstack{N: 2.18} \\

Age under 18
& -1.514*** & -2.97
& -0.828*** & -3.07
& \shortstack{C: -0.815***\\N: -0.815***}
& \shortstack{C: -3.03\\N: -3.03}
& \shortstack{C: -0.453\\N: -0.823***}
& \shortstack{C: -1.57\\N: -3.05} \\

Wet road
& -0.347 & -0.85
& -0.068 & -0.41
& \shortstack{C: -0.190\\N: -0.068}
& \shortstack{C: -0.82\\N: -0.41}
& \shortstack{C: -0.084\\N: -0.084}
& \shortstack{C: -0.85\\N: -0.85} \\

Environment-related factor
& -1.984** & -2.38
& -0.517 & -1.58
& \shortstack{C: -1.071\\N: -0.516}
& \shortstack{C: -1.62\\N: -1.58}
& \shortstack{C: -0.543**\\N: -0.543**}
& \shortstack{C: -1.96\\N: -1.96} \\

Equipment-related factor
& -3.014*** & -5.17
& &
& \shortstack{C: -1.615**}
& \shortstack{C: -2.03}
& \shortstack{C: -0.853}
& \shortstack{C: -1.59} \\

Human-related factor
& 2.031** & 2.36
& &
& \shortstack{C: 1.083}
& \shortstack{C: 1.58}
& \shortstack{C: 0.567}
& \shortstack{C: 1.38} \\

Street/highway location
& &
& -0.247 & -1.30
& \shortstack{N: -0.245}
& \shortstack{N: -1.29}
& \shortstack{N: -0.249}
& \shortstack{N: -1.31} \\

Recreation/sport location
& -2.323*** & -3.35
& &
& \shortstack{C: -1.237*}
& \shortstack{C: -1.81}
& \shortstack{C: -0.652}
& \shortstack{C: -1.53} \\

\addlinespace[2pt]

CPSRMS threshold $\tau_1$
& -2.680*** & -2.65
& &
& -2.672*** & -2.64
& -2.779*** & -2.83 \\

CPSRMS threshold increment $\Delta\tau_2$
& 2.140*** & 6.76
& &
& 2.141*** & 6.77
& 2.136*** & 6.74 \\

NEISS threshold $\tau_1$
& &
& 0.895*** & 4.59
& 0.893*** & 4.58
& 0.894*** & 4.64 \\

NEISS threshold increment $\Delta\tau_2$
& &
& 1.700*** & 13.90
& 1.696*** & 13.91
& 1.696*** & 13.91 \\

\addlinespace[2pt]

Relative scale $k$ (NEISS $=1$)
& &
& &
& 1.868 & 1.33
& 3.548** & 2.15 \\

\midrule
\multicolumn{9}{l}{\textbf{Model fit}} \\

No. of observations
& \multicolumn{2}{c}{278}
& \multicolumn{2}{c}{1,181}
& \multicolumn{2}{c}{1,459}
& \multicolumn{2}{c}{1,459} \\

Final log likelihood
& \multicolumn{2}{c}{-126.391}
& \multicolumn{2}{c}{-751.854}
& \multicolumn{2}{c}{-878.281}
& \multicolumn{2}{c}{-878.417} \\

Akaike Information Criterion (AIC)
& \multicolumn{2}{c}{276.782}
& \multicolumn{2}{c}{1,523.707}
& \multicolumn{2}{c}{1,798.561}
& \multicolumn{2}{c}{1,794.833} \\

Bayesian Information Criterion (BIC)
& \multicolumn{2}{c}{320.313}
& \multicolumn{2}{c}{1,574.448}
& \multicolumn{2}{c}{1,909.557}
& \multicolumn{2}{c}{1,895.258} \\

\bottomrule
\end{tabular*}

\vspace{2pt}

\parbox{\textwidth}{\footnotesize
\textit{Note:} Robust t-statistics are reported.
***, **, and * indicate significance at the
1\%, 5\%, and 10\% levels, respectively.
C and N denote CPSRMS and NEISS.
}

\endgroup
\end{table}

Model 1 identifies several significant associations in CPSRMS.
Human-related factors are positively associated with injury severity
($\beta=2.031$, $p<0.05$), whereas equipment-related
($\beta=-3.014$, $p<0.01$) and environment-related factors
($\beta=-1.984$, $p<0.05$) are negatively associated with severity.
Riders under age 18 ($\beta=-1.514$, $p<0.01$), spring incidents
($\beta=-0.978$, $p<0.05$), and recreation/sport locations
($\beta=-2.323$, $p<0.01$) are also associated with lower
severity categories. Other retained predictors are not
statistically significant.

Model 2 reveals a different pattern in NEISS, where demographic
characteristics are more prominent. Riders under age 18
($\beta=-0.828$, $p<0.01$) and female riders
($\beta=-0.403$, $p<0.05$) are associated with lower
treatment severity, whereas riders above age 59
($\beta=0.395$, $p<0.05$) are associated with higher
severity. The remaining predictors are not statistically
significant.

Model 3 estimates predominantly source-specific coefficients
while sharing the age-under-18 and age-18--24 coefficients
across datasets. The shared age-under-18 coefficient is
negative and significant ($\beta=-0.815$, $p<0.01$),
whereas the age-18--24 coefficient is not significant.
Among source-specific effects, equipment-related factors
($\beta=-1.615$, $p<0.05$) and recreation/sport locations
($\beta=-1.237$, $p<0.10$) are negatively associated
with severity in CPSRMS. In NEISS, age above 59
($\beta=0.392$, $p<0.05$) and female gender
($\beta=-0.404$, $p<0.05$) remain significant.

Model 4 introduces a more parsimonious partially unified
specification. The shared environment-related coefficient
is negative and significant ($\beta=-0.543$, $p<0.05$),
whereas the shared age-18--24 and wet-road coefficients
are not significant. Age under 18 is negatively associated
with severity in NEISS ($\beta=-0.823$, $p<0.01$)
but not in CPSRMS. Within NEISS, age above 59
($\beta=0.400$, $p<0.05$) and female gender
($\beta=-0.402$, $p<0.05$) remain significant.
Other source-specific coefficients are not significant.

The estimated CPSRMS relative scale is $k=1.868$ in
Model 3 and $k=3.548$ in Model 4, with only the latter
significantly different from one ($p<0.05$). Although
Model 3 has a slightly higher log likelihood, Model 4
achieves lower AIC and BIC values with fewer parameters.
Likelihood-ratio tests do not reject equality of the
age-under-18 coefficients ($LR=0.740$, $df=1$, $p=0.390$)
or the additional age-18--24 sharing restriction
($LR=0.072$, $df=1$, $p=0.788$). However, equality
of the source-specific thresholds is rejected
($LR=21.040$, $df=2$, $p<0.001$), supporting
separate severity thresholds across datasets.

Overall, CPSRMS and NEISS provide complementary evidence:
narrative-derived contributing factors are more prominent
in CPSRMS, whereas demographic characteristics are more
prominent in NEISS. The joint models support selected
coefficient-sharing restrictions but not common thresholds.
Accordingly, Models 1 and 2 remain the primary basis for
substantive interpretation, with Model 4 serving as the
more parsimonious supplementary joint specification.

\section{Conclusion}

This study developed an LLM-assisted injury-surveillance framework
to investigate e-bike incidents using two complementary data sources,
CPSRMS and NEISS. By transforming unstructured narratives into
structured contributing-factor and component-level indicators, the
framework integrates contextual information with demographic,
locational, temporal, and weather-derived variables. Injury severity
was harmonized into three ordered categories and examined using
source-specific ordered logit models and two supplementary joint
scaled specifications. In CPSRMS, human-related factors were
associated with higher injury severity, whereas equipment- and
environment-related factors, riders under age 18, spring incidents,
and recreation/sport locations were associated with lower severity.
These negative associations should not be interpreted as causal
protective effects. In NEISS, riders under age 18 and female riders
were associated with lower treatment severity, while riders above
age 59 were associated with higher severity. The joint models
supported selected coefficient-sharing restrictions but rejected
equality of source-specific severity thresholds, underscoring the
importance of preserving differences between the two surveillance
systems.

The framework also demonstrated the potential of LLM-assisted
information extraction for injury surveillance. Manual evaluation
of a stratified 300-case validation sample yielded weighted F1
scores of 0.91 for CPSRMS and 0.86 for NEISS in identifying
mechanical and visibility component associations. Component-level
analysis revealed that CPSRMS captured a broader range of
electrical and mechanical issues, particularly involving drive
systems, batteries, pedals, and wheels/tires, whereas NEISS
more frequently represented wheel/tire and braking-related
involvement. These findings highlight the complementary
information captured by consumer-product incident reports
and emergency-department surveillance records.

The findings suggest several implications for e-bike safety.
Human-related severity associations highlight the relevance
of rider education and behavioral interventions, while
component-related patterns underscore the importance of
battery safety, braking performance, and mechanical reliability.
Demographic differences in NEISS further suggest the value
of age- and gender-sensitive prevention strategies.
However, these implications should be interpreted as
priorities for further investigation rather than causal
effects established by the models.

Several limitations remain. Complete-case estimation,
particularly given substantial missingness in CPSRMS,
may introduce selection bias and limit generalizability.
Manual validation primarily assessed component associations,
whereas the narrative-derived contributing-factor indicators
used in severity modeling require further validation.
The study also did not establish whether zero-shot prompting
outperforms alternative extraction approaches. Differences
in reporting mechanisms, case composition, and severity
definitions further limit direct pooled interpretation,
making the joint models supplementary to the source-specific
estimates. Future research should expand manual validation,
compare alternative extraction methods, investigate missing-data
treatments, and assess robustness across additional years,
jurisdictions, and surveillance systems. Overall, the proposed
framework illustrates how LLM-assisted narrative processing
and econometric modeling can jointly support e-bike injury
surveillance while respecting source-specific differences.

\section*{Declaration of Interest Statement}

The authors declare no conflicts of interest.

\section*{Declaration of Generative AI Use}

OpenAI's GPT-4 was used to analyze e-bike incident narratives and extract structured information. ChatGPT was used for language editing. The authors reviewed all content and take full responsibility for the manuscript.

\section*{Data Availability Statement}

The data were obtained from the U.S. Consumer Product Safety Commission's CPSRMS and NEISS databases and are subject to applicable access policies.


\end{document}